\documentclass{article}

\usepackage[final]{nips_2017}

\usepackage[utf8]{inputenc} 
\usepackage[T1]{fontenc}    
\usepackage{hyperref}       
\usepackage{url}            
\usepackage{booktabs}       
\usepackage{amsfonts}       
\usepackage{nicefrac}       
\usepackage{microtype}      
\usepackage{graphicx}
\usepackage{float}
\usepackage[symbol]{footmisc}

\title{Machine Learning for Exam Triage\thanks{This paper is a written presentation of a project in HackAuton 2018}}

\author{
    Xinyu Guan \\
    Carnegie Mellon University\\
    Pittsburgh, PA 15213 \\
    \texttt{xguan@andrew.cmu.edu} \\
    \And
    Jessica Lee \\
    Carnegie Mellon University\\
    Pittsburgh, PA 15213 \\
    \texttt{jl5@andrew.cmu.edu} \\
    \AND
    Peter Wu \\
    Carnegie Mellon University\\
    Pittsburgh, PA 15213 \\
    \texttt{peterw1@andrew.cmu.edu} \\
    \And
    Yue Wu \\
    Carnegie Mellon University\\
    Pittsburgh, PA 15213 \\
    \texttt{ywu5@andrew.cmu.edu} \\
}

\begin{document}
\maketitle

\begin{abstract}
  In this project we extend the state-of-the-art CheXNet (\citet{rajpurkar2017chexnet}) by making use of the additional non-image features in the dataset. Our model\footnote{Code available at https://github.com/Holmeswww/CheXNetPP} produced  better AUROC scores than the original CheXNet.
\end{abstract}

\section{Introduction}
According to the CDC (\citet{CDCpneumonia}), there are more than 50,000 US deaths annually due to pneumonia. However, the best way currently to diagnose a patient  with pneumonia early is with chest X-rays, which require an expert radiologist's eye to analyze. This results in a fairly long turn around that many hospitals would like to reduce. CheXNet (\citet{rajpurkar2017chexnet}), a machine learning algorithm that can analyze chest X-rays in bulk, was built to help solve this problem. In this paper, we improved slightly upon CheXNet, by allowing the network to take in additional information about the patient outside of the X-Ray.

The paper is structured as follows. In Section \ref{background}, we give some background on CheXNet \citet{rajpurkar2017chexnet}. Section \ref{Arch} explains our neural network architecture. Section \ref{Exp} focuses on the experimental results and Section \ref{Conc} offers a brief discussion.

\section{Background}
\label{background}

The previous state of the art on this data set is CheXNet, which is a 121-layer Densely Connected Convolutional Neural Network (DenseNet) that classified X-Ray images at the accuracy of radiologists. The previous state of the art only outputted a binary classification of whether pneumonia was present. CheXNet improved on that by replacing the last fully connected layer to recognize 14 classes - Atelectasis,Cardiomegaly, Consolidation, Edema, Effusion, Emphysema, Fibrosis, Hernia, Infiltration, Mass, Nodule, Pleural Thickening, Pneumonia, and Pneumothorax. Some limitations of ChexNet were the lack of use of patient history and of lateral images. 

\section{Network Architecture}
\label{Arch}
\begin{figure}[H]
  \centering
  \includegraphics[width=\linewidth]{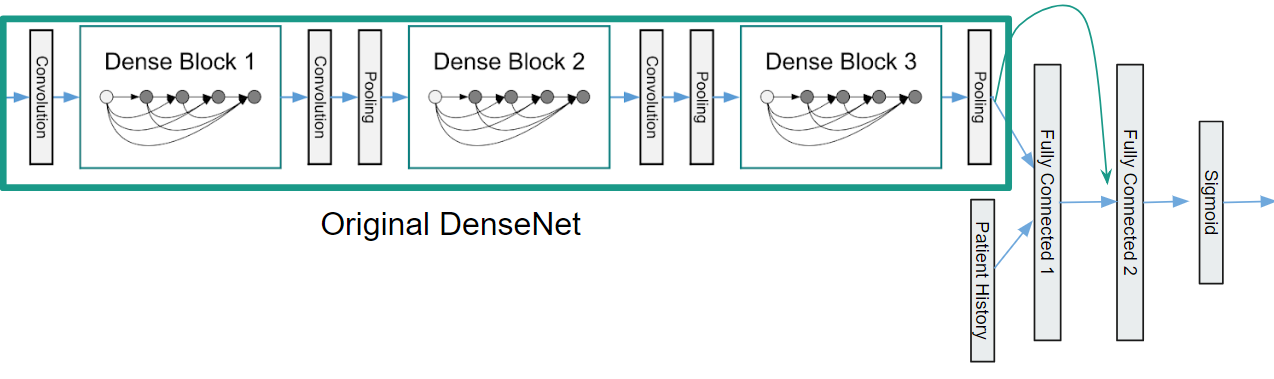}
\end{figure}
Our model uses the original DenseNet (\citet{huang2017densely}) with two fully connected layers (ReLU activation). The skip connection (\citet{he2016deep}) from the last pooling layer of the DenseNet to the second fully connected layer encourages gradient flow, and ensures a worst-case identity map so that our model should perform at least as good as CheXNet (\citet{rajpurkar2017chexnet}).

\section{Experiments}
\label{Exp}

We trained our model \footnote{We based our code on a github implementation of CheXNet: https://github.com/zoogzog/chexnet} on the full ChestX-ray14 dataset, using exactly the same learning parameters as in (\citet{rajpurkar2017chexnet}). Our model was trained with a batch size of 16, a 20:10:70 test:val:train split for 44 epochs, and did not train to convergence.

\begin{table}[H]
  \caption{Performance of our model compared to the reported performance of CheXNet (\citet{rajpurkar2017chexnet}). Values reported are the AUROC scores each model achieved.}
  \label{Comparison of our work against CheXNet}
  \centering
  \begin{tabular}{llll}
    \toprule
        Pathology &	 ChexNet    & Our Model \\
    \midrule
    Atelectasis   &	0.8094 &	0.8328 \\
    Cardiomegaly   &  	0.9248 &	0.9012 \\
    Effusion     &	0.8638 &	0.8911 \\
    Infiltration &	0.7345 &	0.7205 \\
    Mass    &	0.8676 &	0.8814 \\
    Nodule  &	0.7802 &	0.8175 \\
    Pneumonia &	0.768	& 0.7665 \\ 
    Pneumothorax &	0.8887 &	0.9145\\ 
    Consolidation &	0.7901 &	0.8155\\ 
    Edema   &	0.8878	& 0.9138\\
    Emphysema & 	0.9371 &	0.9271 \\
    Fibrosis &	0.8047 &	0.8221 \\ 
    Pleural Thickening &	0.8062	& 0.8110 \\
    Hernia &	0.9164 &	0.9733\\
    \bottomrule
  \end{tabular}
\end{table}

\section{Discussion}
\label{Conc}
In this project, we demonstrated how additional non-image features may benefit the overall performance of the model.
Future research can focus on training with higher resolution, training with different depth, applying model compression, and hyperparameter optimization. Furthermore, as our model did not reach convergence at 44 epochs, we believe it may perform even better when it reaches convergence. 

\subsubsection*{Acknowledgments}

We would like to thank the Pittsburgh Supercomputing Center for providing computation resources, and the Auton Lab for providing such a wonderful opportunity.

\bibliographystyle{plainnat}
\bibliography{cite}






\end{document}